\documentclass{article}
\usepackage{acra}
\usepackage{amsmath, amssymb}
\usepackage{graphicx}
\usepackage[utf8]{inputenc}

\title{Kiwifruit detection in challenging conditions}
\author{Mahla Nejati$^{1}$$^{*}$, Nicky Penhall$^{1}$, Henry Williams$^{1}$, Jamie Bell$^{1}$,\\ JongYoon Lim$^{1}$, Ho Seok Ahn$^{1}$, Bruce MacDonald$^{1}$ \\ $^{1}$The Centre for Automation and Robotic Engineering Science, \\the University of Auckland, Auckland, New Zealand \\ 
$^{*}$mnej691@aucklanduni.ac.nz}

\begin{document}

\maketitle

\begin{abstract}
        Accurate and reliable kiwifruit detection is one of the biggest challenges in developing a selective fruit harvesting robot. 
        The vision system of an orchard robot faces difficulties such as dynamic lighting conditions and fruit occlusions.
        This paper presents a semantic segmentation approach with two novel image prepossessing techniques designed to detect kiwifruit under the harsh lighting conditions found in the canopy.  
        The performance of the presented system is evaluated on a 3D real-world image set of kiwifruit under different lighting conditions (typical, glare, and overexposed).
        Alone the semantic segmentation approach achieves an F$_1$score of 0.82 on the typical lighting image set, but struggles with harsh lighting with an F$_1$score of 0.13. Utilising the prepossessing techniques the vision system under harsh lighting improves to an F$_1$score 0.42. To address the fruit occlusion challenge, the overall approach was found to be capable of detecting 87.0\% of non-occluded and 30.0\% of occluded kiwifruit across all lighting conditions.
\end{abstract}

\section{Introduction}
    
    Kiwifruit growers face shortages of seasonal workers, and the high cost of human labour is a major impediment to this target. In addition, picking kiwifruit is a tedious and repetitive job, and workers need to carry a heavy picking bag which increases the likelihood of back strain and musculoskeletal problems. From the perspective of the orchard owner, they need to train workers for picking, and health and safety in the orchard. To prevent hazards in the orchard, they must inspect orchards frequently \cite{Safety2016} which adds additional costs.
    
    Given the above problems, the kiwifruit industry would benefit from automation, especially around the harvesting. An autonomous kiwifruit harvesting robot would: compensate for the lack of workers, reduce labour costs, and increase fruit quality. The work presented here is a part of a wider project to develop robots that drive through kiwifruit orchards, while pollinating and harvesting kiwifruit \cite{Williams2019,Williams2019H2,Williams2019H1,Barnett2017}.
    
\begin{figure}[!htbp]
      \centering
        \parbox{3in}
            {\includegraphics[width=3in, height=5.5cm]{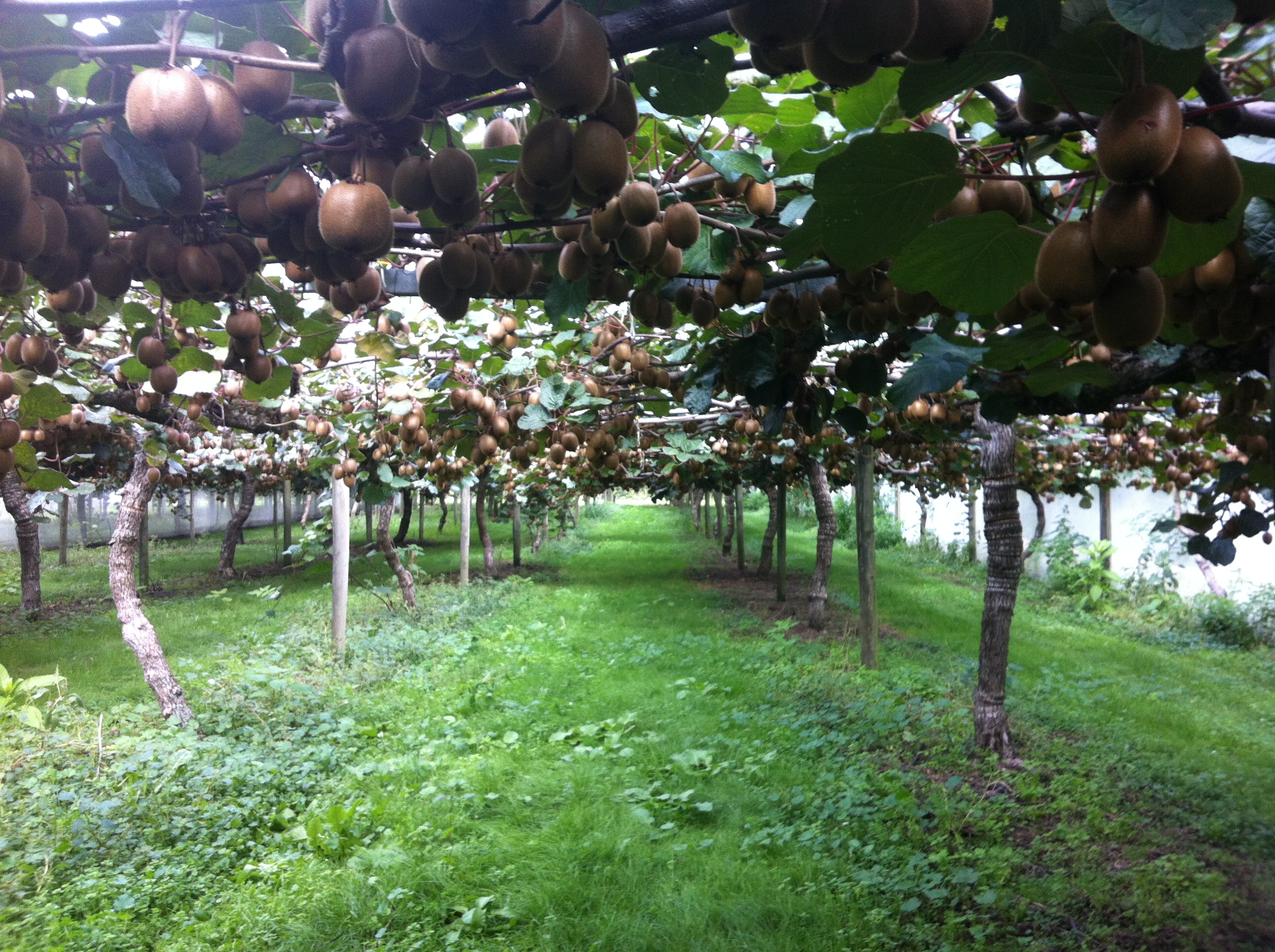}}
      \caption{A kiwifruit orchard}
      \label{fig:kiwifruitorchard}
   \end{figure}

    Kiwifruit is a delicate fruit and needs to be picked gently. The kiwifruit orchard has a pergola structure which kiwifruit grows downward. An example of kiwifruit orchard is shown in figure \ref{fig:kiwifruitorchard}. Bulk harvesting methods, like shakers, are not suitable if kiwifruit is to be sold in fresh markets as they may cause bruising \cite{Li2011}. Therefore, a selective fruit harvesting robot is required. 
    
    Automatic fruit detection is a key component in a selective fruit harvesting robot. Working on automatic detection of fruit started from 1968 \cite{Brown1968}, despite this, no commercial selective harvesting robot is available yet, and the performance is insufficient to compete with manual labours. 
    
    The main reasons for the lack of commercial harvesting robots and poor performance are due to the low accuracy of fruit recognition methods and the inefficiency of localization methods \cite{Zhao2016}. In 1968, Brown et al. \cite{Brown1968} reported fruit/flower occlusion and uneven illumination conditions as critical problems. Still, after 50 years, many researchers today cite these as major challenges to produce a commercially viable autonomous kiwifruit harvester. 
    
    The aim of the paper is presenting the performance of kiwifruit detection using convolutional neural network in unpredictable lighting conditions and propose a prepossessing to improve the performance. According to detection challenges, two datasets including 295 images were collected to evaluate the performance of proposed detection method which are kiwifruit occlusion and variable lighting conditions. The remainder of the paper is organised as follow: section \ref{sec:RelatedWork} describes the related work for kiwifruit detection and solutions to overcome various environmental conditions. In section  \ref{sec:DetectionMethod}, the detection approach is presented, which uses a fully convolutional neural network with a proposed preprocessing method. The experimental setup and our findings are presented in section  \ref{sec:Results}. Conclusions are drawn in section \ref{sec:Discussion} with future directions.

\section{Related Work}\label{sec:RelatedWork}
    Existing methods employed for fruit detection are classified into two separate groups: hand-engineered feature extraction, and machine learning methods. Popular hand-engineered features are: colour, shape, texture and fusion of features \cite{Gongal2015}. The use of a threshold value on a hand-engineered feature is one of the basic approaches for identifying a piece of fruit. Fu et al. \cite{Fu2015} applied Otsu's method to kiwifruit images, on the R-G colour space. The result was a binary image where morphological operations are used to reduce the noise below a given detection threshold. Via this binary image, edge detection and Circle Hough Transform (CHT) method were shown to be useful for extracting ellipses that indicated kiwifruit in the image. The author showed, during evaluation, that at the pixel-level the detection rate was 88\% correct using images with 12 different illumination levels. In subsequent work, the authors \cite{Fu2017} tried to detect the calyx of kiwifruit. After finding the boundary of the kiwifruit (using the above methods), the calyx was extracted using the Otsu's method based on the V channel in HSV colour space. Correct detection rate based on pixel-wise evaluation in different illumination levels was reported as 94\%. Despite the high detection rate, the proposed method depends on the cluster size and calyx location in the fruit. Both studies captured kiwifruit data at night time. The moisture on the skin of the kiwifruit causes some challenges in post-processing harvesting procedure due to this issue kiwifruit were not harvested at night time. Therefore, the proposed kiwifruit detection should be able to work in the daytime and under uneven lighting conditions.
    
    The problem with the hand-engineered feature based techniques is that they can work reliably only under controlled conditions. In contrast, learning methods are robust to dynamic conditions such as lighting, camera viewpoint, fruit occlusion and fruit size. Wijethunga et al. \cite{Wijethunga2009} used a Self-Organizing Map (SOM) model to detect kiwifruit. The normalised pixel data in L*a*b* colour space was used for training the model, and the author concluded that its performance was better compared to using the original data - although the author has not quantified the result. 
    
    Zhan et al. \cite{ZhanW} applied six weak classifiers on RGB, HSI and La*b* colour spaces by using an Adaboost algorithm to detect kiwifruit. The author evaluated the method on 208 kiwifruit pixels and 477 background pixels. It was reported that there was 97\% accuracy at the pixel-level, with a 7\% false detection rate. The performance of the proposed algorithm is quite high. However, it was tested on a small dataset and needed some post-processing to discover the fruit. Moreover, one of the important indicators to evaluate a detection method for a real-time robot is the processing time which was not revealed. 
    
    Regarding the variations in illumination conditions, some researchers have attempted to solve this problem by restricting the environment or preprocessing the data. Bargoti et al. \cite{Bargoti2016} suggested adding the azimuth of the sun as input data to a multilayer perceptron. The author showed that the performance improved negligibly when the angle of the sun was represented by two values on the input. The author in his next research \cite{Bargoti2017a} added the azimuth of the sun to a Convolutional neural network and achieved the same conclusion. It seems more characteristics should be used in presenting the sun's status rather than two values. 
    
    Some researchers implemented preprocessing operations to overcome this issue. For example, Zemmour et al. \cite{Zemmour2017} divided the image into sub-images with a homogeneous lighting conditions and then categorized them into groups based on their level of illumination: low, mid or high. This method can be applied to any images and does not require any external hardware, although it is time-consuming. Another approach is using exposure fusion to capture images \cite{Silwal2017}. It means images with different exposure times were captured from the same scene. Then, the best region was computed for each image, and those regions were then fused into the final image. Although this method will reduce glare or dark images, the capturing of the image will be slow, and it is not suitable for real-time application. 
    
    Another solution is changing the capturing environment. For example, capsicum detection in \cite{Sa2016a} was conducted in a commercial field under controlled lighting conditions. Jimenez et al. \cite{A.R.Jimenez2000} suggested that using artificial lighting can reduce shadows caused by sunlight. Another suggestion, for having consistent lighting conditions, was capturing images at nighttime \cite{Fu2015,Fu2017,Wijethunga2009,Wanga2013} or on an overcast day \cite{Dias2018}.
    
    Another approach was using a plain background (e.g. sheet/board/cover) behind the tree to achieve robust detection results \cite{Silwal2017,Nguyen2016,Baeten2008}. By using this technique, researchers achieved good results, especially on apples. However, using a plain background is related to the structure of the orchard. Hence this method can not be used on the kiwifruit orchard. Reducing the lighting to a single source, such as a camera flash, at night time would be useful. However, it would not be useful for all situations in the daytime. In a kiwifruit orchard, the camera faces the sun, hence adding artificial lighting, in theory, may help to capture a better image. Taking images with uneven illumination conditions is not avoidable.
    
    With regard to the occlusion challenge, it is related to the production environment and the type of orchard and fruit. In New Zealand, most kiwifruit orchards are built on a pergola structure which makes kiwifruit hang downward in the canopy. The pergola structure is built by using posts, supporting beams, and supporting wires. 
    
    Occlusion --- from the perspective of the camera --- usually occurs when the fruit is occluded by at least one obstacle such as a: leaf, fruit, branch, wire, supporting beam or post. The density of kiwifruit, foliage coverage and thinning method effect on the degree of occlusion. Classifying the fruit data into occluded and non-occluded classes is time-consuming and subjective. Despite these difficulties, it can be a good reference to compare the performance of the fruit detection method. Researchers reported fruit occlusion as a challenge for hand-engineered feature based techniques. For example, one of the common approaches for detecting circular fruit is using CHT which is not able to detect occluded fruit \cite{Hiroma2002}.

\section{Detection Method}\label{sec:DetectionMethod}

    This section presents the Fully Convolutional Network (FCN) for kiwifruit calyx detection and introduces the transfer learning and preprocessing techniques related to improving performance under challenging lighting conditions.
    
    The key requirements for the detection method were that the boundaries of kiwifruit and obstructions, such as wires and branches in the canopy, should be detected. Bounding boxes do not define the boundaries of thin obstructions well when the obstructions are at an angle to the rows and columns of the image. For example, in an image of the kiwifruit canopy, a thin wire that starts at one corner of an image and ends at the opposite corner would have a ground truth bounding box of the whole image, even though the wire only takes up a small area of the image and the canopy shown in the image. Instead, it was decided to use a semantic segmentation neural network which can identify the pixels associated with kiwifruit, branches and wires. The information about the existence of wires and branches helps the arm to do not try to pick the kiwifruit close to obstacles or approaching those kiwifruits with a certain angle and it will prevent the gripper to getting stuck in obstacles. Kiwifruit grows in clusters so recognizing individual fruit in a cluster was a challenge. Hence, instead of detecting the skin of the kiwifruit, only the calyx area was identified.
    
\begin{figure}[!htbp]
      \centering
        \parbox{3in}
            {\includegraphics[width=3in, height=4cm]{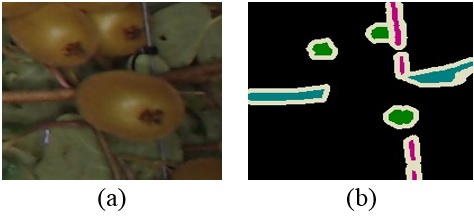}}
      \caption{An example of training data which includes (a) input image (b) annotated image green, cyan and pink colour indicates calyx of the kiwifruit, branch and wire, respectively}
      \label{label1}
   \end{figure}

    After experimenting with different instances of the FCN for semantic segmentation, the architecture selected for use was FCN-8s \cite{Long2015}. In order to train FCN-8s, 63 images with 113 kiwifruits were labelled manually with kiwifruit calyx, branch and wire classes. In labelling for the training dataset, there was not any classification on calyxes such as occluded and non-occluded and all calyxes were labelled as the calyx. The dimensions of the RGB input and indexed label images were 1920$\times$1200 pixels. The dataset was divided into 48 training and 15 validation images. The inputs for training were cropped to 200$\times$200 pixels from the full sized images and label images in order to limit the memory used during training. An example input image and annotation image pair are shown in figure \ref{label1}. 
    
    Images were taken in 2015 and 2016 from a range of kiwifruit orchards at different times of the day and with different weather. A tool was implemented for labelling segments, with which the user selects the boundary, and the class of the segment and the tool fills segments with a colour that has been assigned to the class. Pixel-wise annotation is a time-consuming process, and it took around 30 minutes for labelling three classes per image.
    
    In transfer learning, an adapted version of VGG16 net \cite{Simonyan2014} equivalent to 19 convolutional layers was used, with feature windows of eight pixels. The model with this configuration is called FCN-8s and was provided for Caffe via GitHub\footnote{https://github.com/shelhamer/fcn.berkeleyvision.org}. The PASCAL VOC dataset was used for pre-training, so the number of classes represented by the model was 21. The PASCAL dataset has 20 classes (21 including the borders), and then the final label was ignored because it represented borders between classes.
    
\begin{figure}[!htbp]
      \centering
        \parbox{3in}
            {\includegraphics[width=3in, height=5cm]{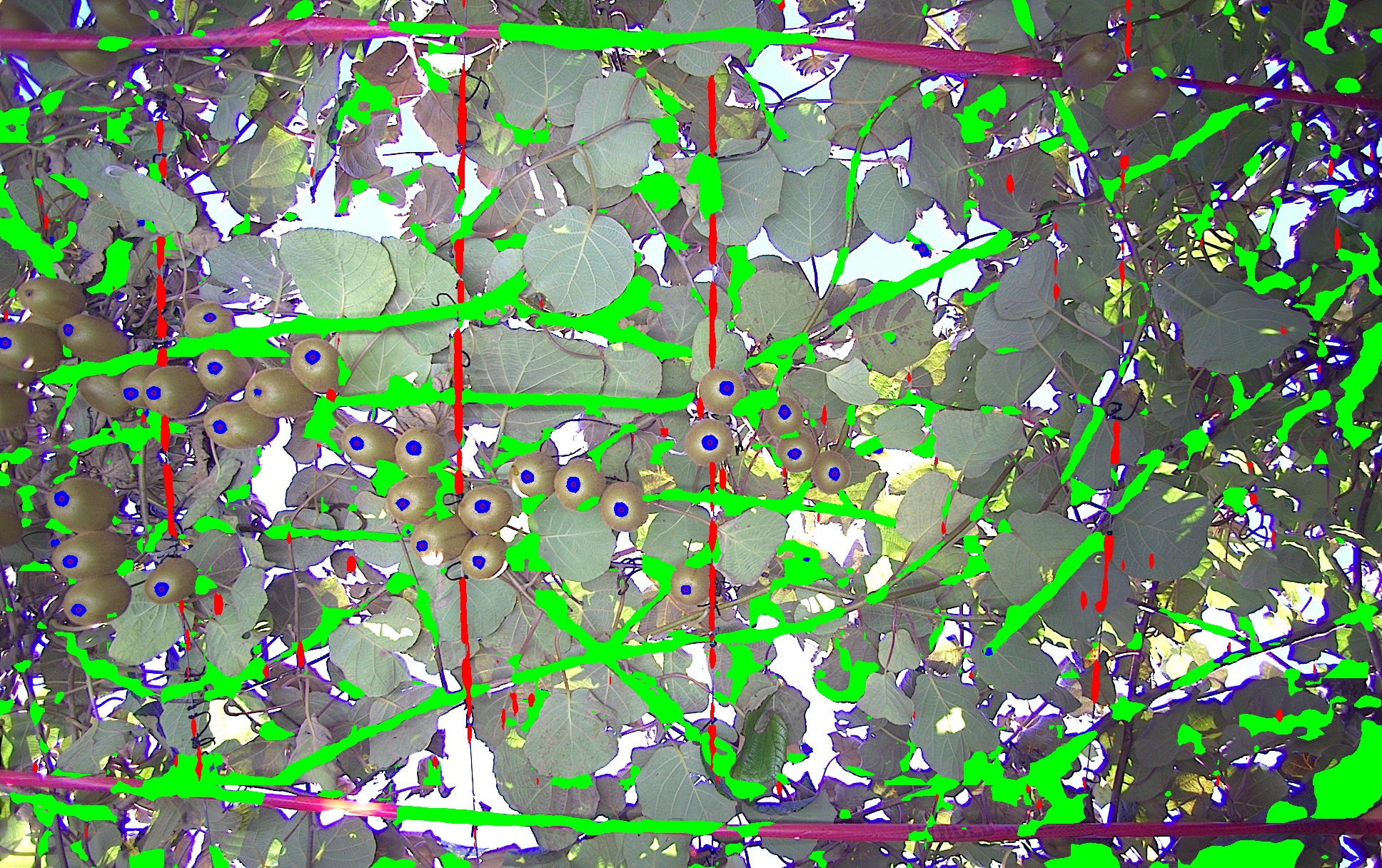}}
      \caption{The output of applying FCN-8s on an image, pixels with blue, red and green colours show calyx, wire and branches, respectively}
      \label{label2}
   \end{figure}

    NVIDIA DIGITS (version 5.1-dev) was used for training the model using the NVIDIA Caffe back-end (version 0.15.14) on an Nvidia GTX-1070 8GB graphics card. The maximum image size was limited to 200$\times$200 pixels due to the hardware constraints. The result was achieved after convergence at a 0.0002 learning rate with 1000 epochs. Instead of a standard stochastic gradient descent procedure, the Adam algorithm was used for updating network weights. 
    The output of the model was a $h\times w\times d$ matrix which $h\times w$ was the image size, and $d$ value refered to the number of classes (21) in PASCAL VOC dataset. The output matrix showed the probability of being the specified class for each pixel. Applying a maximum function to values of $d$ for each pixel gave us the index of the class that the pixel belonged. An example of the output is presented in figure \ref{label2}. 
    
    FCN-8s is a pixel-based method and indicated segments show the result of the detection. In order to identify and extract calyx regions, blob detection was used to determine the centre and the radius of the calyx. The standard OpenCV blob detection was used for extracting blobs. The Minimum thresholds were set to the area size (150 pixels) and the circularity (0.5) parameters. 
    
\begin{figure}[!htbp]
      \centering
        \parbox{3in}
            {\includegraphics[width=3in, height=4cm]{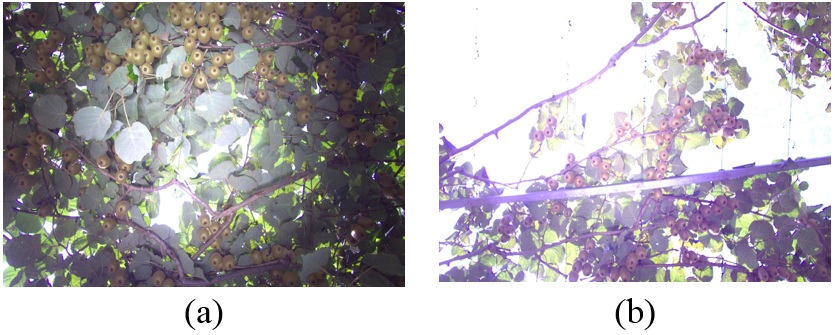}}
      \caption{An example of kiwifruit images (a) a typical image (b) an overexposed image}
      \label{label3}
   \end{figure}

    Due to the pergola structure of kiwifruit orchards, the camera was facing the sun. Therefore, a wide range of images with various lighting conditions can be captured. Images were classified to typical, overexposed and glare images based on lighting conditions. An example of these images is shown in figure \ref{label3}.
    
    The maximum image size that the trained model can be run on is $500\times500$ pixels. It is larger than the input training images due to the inference not needing any backward and forward pass. Therefore, the input image with the resolution of $1936\times1216$ was divided to $500\times500$ sub-images. Each sub-image had an overlap with neighbour windows to reduce the impact of objects sitting close to the edge of a crop being wrongly detected. This overlap was set to 20\% of $\frac{width}{height}$. To handle overlap areas, where there may be conflicting results on what the pixel was, the maximum confidence and class were chosen between the crops.
    
    The model was trained using typical images. Therefore, to overcome unexpected lighting conditions, preprocessing was applied to images. In overexposed images, a high number of pixels was saturated. Besides, the distribution of light was not equal, and sometimes a region of the image was dark, and the fruit was hard to see. 
    
\begin{figure}[!htbp]
        \centering
            \parbox{3in}
                {\includegraphics[width=3in, height=5cm]{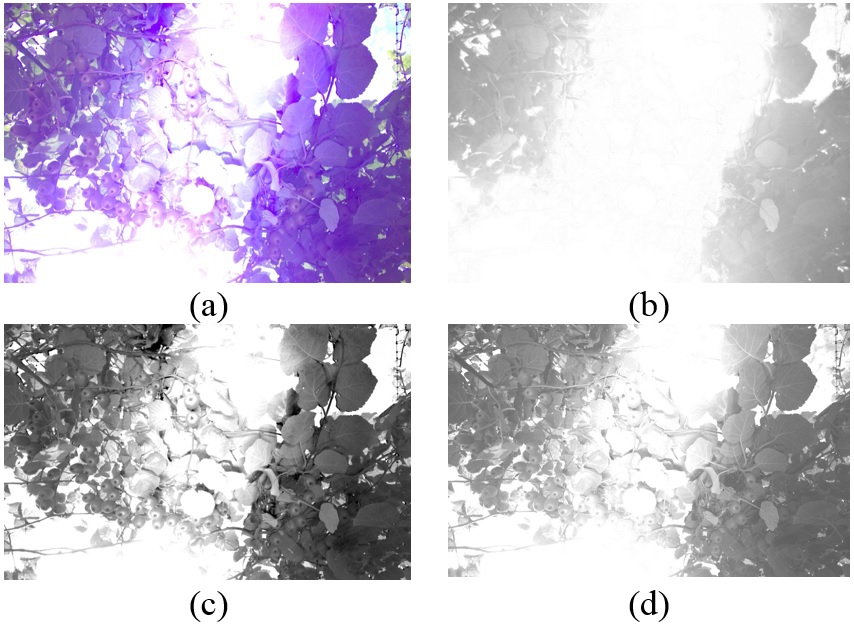}}
        \caption{An example of a (a) glare image (b) its blue channel  (c) its green channel  (d) its red channel}
        \label{label4}
    \end{figure}

    The glare image is distinguished visually by a cover of purple colour on the majority part of the image. This means the image has a high number of saturated pixels in the blue channel (figure \ref{label4}(a)). 
    
    In order to automatic classify the overexposed and glare image a definition based on the number of saturated pixels are defined. An image is overexposed if the number of saturated pixel in R ($R_s$), G ($G_s$), and B ($B_s$) channels are more than a quarter of the image size ($S$), as follows:
    \begin{equation}
        {\frac {B_s \ and \ G_s \ and \ R_s} {S} \ge 0.25 }
    \end{equation}
    An image is defined glare if the image in overexposed and the number of overexposed pixels in channel blue is more than half of the image size.
    \begin{equation}
    \mathit{Glare = \frac {B_s \ and \ G_s \ and \ R_s} {S} \ge 0.25 \ and \ \frac {B_s} {S} \ge 0.5}
    \end{equation}

    Regarding the preprocessing technique for an overexposed image, the image was converted from RGB channel to YCbCr, and then histogram equalization (HE) was applied on the Y channel and converted back to the RGB channel. Then, the image was split into 500$\times$500 and the FCN-8s applied to each part. 
    
    A glare image is shown in figure \ref{label4} in different channels. You can see, the green channel has more information, and the blue channel is affected by the sun. Hence, in this kind of image, the blue channel was exchanged for the green channel. The histogram equalisation was applied to each 500$\times$500 block, due to the lighting conditions being more dynamic when compared to the overexposed images. At the end, FCN-8s was applied on each block.    
    
   \section{Results}\label{sec:Results}
    Two main challenges of fruit detection are various lighting conditions and the fruit occlusion \cite{A.R.Jimenez2000}. Therefore, a suitable kiwifruit detection method needs to work reliably and fast in both conditions. In this section, first, the dataset is described and then the performance of the proposed method is evaluated in different lighting conditions and with occluded fruit. Performance is based on existing evaluation metrics.

    \subsection{Dataset}
    Images were captured by two Basler ac1920-40uc USB 3.0 colour camera with Kowa lens LM6HC (similar setting to \cite{Nejati2019}). In order to have consistent lighting in images, a pair of light bars were mounted alongside the cameras. Images of the dataset were collected during daylight hours at a kiwifruit orchard in Tauranga on the 23rd of March 2017 by a test rig. The dataset consisted of two different sub-datasets for testing the performance of the detection method in various lighting conditions and the kiwifruit occlusion. 
    
\begin{figure}[!htbp]
        \centering
            \parbox{3in}
                {\includegraphics[width=3in, height=4.5cm]{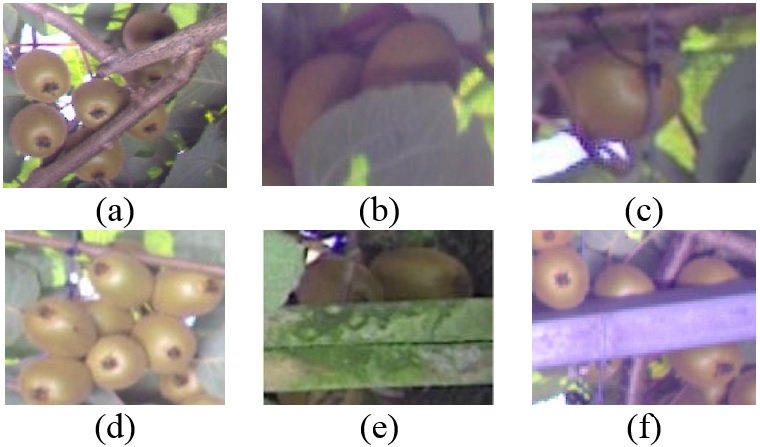}}
        \caption{An example of occluded calyx by a (a) branch (b) leaf (c) wire (d) fruit (e) post (f) supporting beam}
        \label{label5}
    \end{figure}

   In the kiwifruit occlusion dataset, 50 images were selected randomly which included natural lighting conditions. The definition of occlusion is when half of the calyx is occluded by at least one of the obstacles. The number of visible calyx and occluded ones were counted in each image manually. Figure \ref{label5} depicts examples of objects that could occlude a calyx.
   
    Then, the proposed detection method without preprocessing was applied to images, and the performance on occluded and non-occluded calyxes was compared.
    
    To cover the various lighting conditions, another dataset was selected randomly from the captured images. The dataset was divided into three classes which were typical, overexposed and glare images. The definition of these terms was discussed in the prior section. Then, from each class, some images were selected randomly and an amateur image annotator hand-labelled the calyxes using an object annotation toolbox, and an expert image annotator reviewed all images. 
    
    In order to compare datasets, two parameters were considered which were the number of images and density of calyxes is shown in table \ref{table1}. The density of calyxes is calculated, as follows:
    \begin{equation}
       \mathit{ density = \frac {Total\ number\ of\ calyxes} {Total\ number\ of\ images} }
    \end{equation}
    Density is the average number of calyxes per image in a dataset. Generally, a high density causes poor visibility and more occlusion in an image.
    
    \begin{table}[h]
        \caption{The number of images and density of calyxes in datasets}
        \label{table1}
        \begin{center}
            \begin{tabular}{|c||c|c|}
                \hline
                Dataset & Number of images & Density\\
                \hline
                Kiwifruit occlusion & 50 & 62.7\\
                \hline
                Typical lighting & 121 & 59.29\\
                \hline
                Overexposed lighting & 63 & 53.6\\
                \hline
                Glare lighting & 61 & 82.8\\
                \hline
            \end{tabular}
        \end{center}
    \end{table}
    \subsection{Indicators For Evaluation}
    Four indicators were used for comparing the proposed detection method on different datasets includes recall, precision, F$_1$score and processing time.
    
    Recall is the number of the calyxes that detected correctly ($TP$) over the total number of calyxes in an image ($TP+FN$) and computed as follows: 
    \begin{equation}
        \mathit{Recall = \frac {TP} {TP+FN} }
    \end{equation}
    Precision shows the reliability of the method. It is the number of the calyxes that detected correctly over the total number of calyxes that detected in an images ($TP+FP$) and calculated as: 
    \begin{equation}
        \mathit{Precision = \frac {TP} {TP+FP} }
    \end{equation}
    F$_1$score gives a better idea of the method performance. It is the  combination of the precision and recall via the following equation:
    \begin{equation}
       \mathit{ F_1score = 2 \cdot \frac {Precision \cdot Recall} {Precision + Recall} }
    \end{equation}
    
    \subsection{Kiwifruit Occlusion Dataset}
\begin{figure}[!htbp]
        \centering
            \parbox{3in}
                {\includegraphics[width=3in, height=4cm]{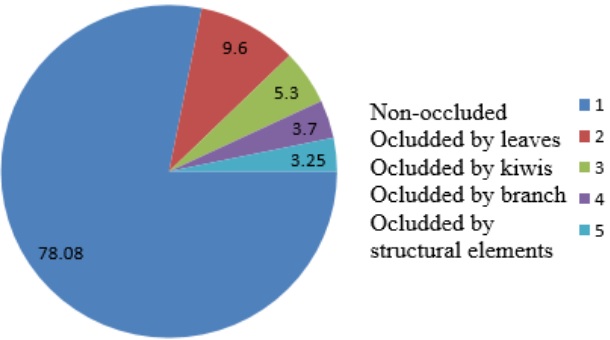}}
        \caption{The distribution of the status of the calyx of the kiwifruit}
        \label{label7}
    \end{figure}

    First, the proposed detection method was applied to the dataset. The performance of the detection method was measured by three indicators which were recall, precision, and F$_1$score of the overall visible calyxes (table \ref{table2}). The F$_1$score was high, so the method would be promising.
    \begin{table}[h]
        \caption{The overall performance of the detection method on the kiwifruit occlusion dataset}
        \label{table2}
        \begin{center}
            \begin{tabular}{|c||c|c|}
                \hline
                Recall & Precision & F$_1$score\\
                \hline
                0.75 & 0.92 & 0.82\\
                \hline
            \end{tabular}
        \end{center}
    \end{table}
    The dataset was split up into the occluded and non-occluded calyxes due to the performance of the detection method being different between them. The number of non-occluded calyxes over the total number of calyxes in the dataset were counted manually and presented as the percentage of non-occluded calyxes in figure \ref{label7}. According to figure \ref{label7}, the percentage of non-occluded calyxes is 78.08\% which shows by the blue colour in the pie chart; the percentage of occluded being 21.92\%. Because the goal was showing the effect of the calyx occlusion on detection, the detection performance was measured on the occluded and non-occluded calyxes separately.
    
    Table \ref{table:Occlusion} shows the recall of the detection method on the non-occluded calyxes was 0.87, while on the occluded calyxes was 0.3. This means if the calyx was fully visible, the method could detect accurately with a high probability. 
    
\begin{figure}[!htbp]
        \centering
            \parbox{3in}
                {\includegraphics[width=3in, height=3cm]{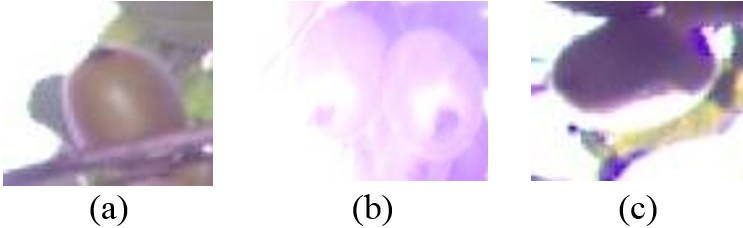}}
        \caption{An example of (a) tilted (b) glare (c) overexposed kiwifruit}
        \label{label6}
    \end{figure}

    \begin{table}[h]
        \caption{The performance of the detection method on occluded and non-occluded kiwifruit}
        \label{table:Occlusion}
        \begin{center}
            \begin{tabular}{|c||c|c|}
                \hline
                 Detection performance & Non-occluded & Occluded\\
                \hline
                Recall & 0.87 & 0.30 \\
                \hline
            \end{tabular}
        \end{center}
    \end{table}
    According to table \ref{table:Occlusion}, the reason that 0.23 of non-occluded calyxes were not detected was due to the various lighting conditions and the complexity of the orchard, with some kiwifruit being tilted. An example of these conditions is shown in figure \ref{label6}.
    
    In order to discover which object has more influence on occlusion, the contribution of each object in calyx occlusion was counted, this is illustrated in figure \ref{label7}. Among all objects which caused occlusion, leaves had occluded 9.6\% of calyxes which was the highest rate. The structural elements had the lowest percentage due to there were fewer of them in the orchard compared to other objects.

    \subsection{Various Lighting  Conditions}
    The results shown in the fruit occlusion dataset subsection was conducted without applying any preprocessing. In this section, related to the category of lighting conditions, the relevant preprocessing was applied to images. Automatic counting was utilised to evaluate the detection method instead of manually counting. The result of the detection method was presented by the centre and the size of a blob which indicates the calyx region. However, the ground truth was annotated by a bounding box around the calyx area. For evaluation, finding the correspondence between each calyx in the predicted and ground truth set was required. Therefore, the Euclidean distance (in pixel units) between each calyx in the two sets had been computed. All calyxes which had a distance greater than a defined threshold was assigned with the highest cost. In our experiments, the threshold was set as 20 pixels.
    
    Then, the Hungarian method \cite{Robinson1955} was used as an optimizer to assign each calyx in the detection set with the lowest cost matched calyx in the ground truth set. Therefore, true positives are when the predicted calyxes have a correspondent in the ground truth set. The remaining calyxes in the predicted and ground truth set were false positives and true negatives, respectively. 
    
    To compare the influence of the preprocessing technique, the detection method was applied to three datasets without any preprocessing. You can see results in table \ref{table3}; the method has the lowest recall on glare images.
    
    \begin{table}[h]
        \caption{The comparison between the performance of the detection method (FCN-8s) and the proposed method (preprocessing and FCN-8s) on various lighting conditions}
        \label{table3}
        \begin{center}
            \begin{tabular}{|c||c|c|c|}
                \hline
                Method/Dataset & Recall &Precision & F$_1$score\\
                \hline
                FCN/Typical & 0.74 & 0.92 & 0.82\\
                \hline
                FCN/Overexp & 0.41 & 0.7 & 0.52\\
                \hline
                Proposed/Overexp & 0.45 & 0.7 & 0.55\\
                \hline
                FCN/Glare & 0.07 & 0.64 & 0.13\\
                \hline
                Proposed/Glare & 0.30 & 0.71 & 0.42\\
                \hline
            \end{tabular}
        \end{center}
    \end{table}

    The precision in all conditions was quite acceptable. It means there will be a small number of false positives. However, the recall was low in glare images. After applying the preprocessing to overexposed and glare images, the result had improved. The result is shown in table \ref{table3}. The improvement in recall for overexposed images is negligible, while it is a significant improvement in the glare images. 
    
    \subsection{Processing Time}
    
    The trained model was deployed on a PC that has Intel Xeon(R), 64-bit 2.40GHz $\times$16 CPUs, Quadro P6000/PCIe/ SSE2 GPU and 32 GiB RAM on an Ubuntu 16.04 Linux system. The processing time for an image includes: cropping, preprocessing, detection, and merging on an image with 1936$\times$1216 px resolution. The average processing time of these image process without the preprocessing part was 1.5 s. The comparison of processing time is shown in table \ref{table5}. 
    
    \begin{table}[h]
        \caption{Average processing time for different lighting conditions}
        \label{table5}
        \begin{center}
            \begin{tabular}{|c||c|}
                \hline
                Images & Average processing time (s) \\
                \hline
                Typical & 1.51 \\
                \hline
                Overexposed & 1.53\\
                \hline
                Glare & 1.59\\
                \hline
            \end{tabular}
        \end{center}
    \end{table}
    
    The average processing time of preprocessing for glare images is 90ms and for overexposed images is 20ms which is negligible. Moreover, the processing time of blob detection is related to the number of detected calyxes.

    \section{Discussion}\label{sec:Discussion}
    
    The result of the detection method depends on the degree of occlusion and lighting conditions. The performance of the method was discussed by showing F$_1$score and processing time. 
    Regarding the calyx occlusion, the level of occlusion is related to the thinning method, orchard structure, foliage coverage and density of kiwifruit. Therefore, the performance of the detection method will be different from orchard to orchard and from year to year. Moreover, there are different types of objects which cause calyx occlusion. According to figure \ref{label7}, 9.6\% of calyxes are occluded by leaves, which the user can remove by hand or using an air blowing system \cite{Dobrusin1992}. Wires and branches occluded around 7\% of calyxes. In the real world, these kiwifruits are obstructed and hard for a robotic arm to pick so maybe they would be detected and picked in the second trial with another method. Moreover, 5.2\% of calyxes are overlapped by other calyxes, and they could become visible when other kiwifruit are picked. 
    
    Among current research that has been done on kiwifruit detection, Fu et al. \cite{Fu2015,Fu2017} captured kiwifruit data at night time to control the lighting conditions, and no information about false positives was released. Only Zhan et al. \cite{ZhanW} used daytime data and presented recall and precision. Although, the evaluation was done at the pixel level and it could not guarantee that the method can detect individual kiwifruit. In all lighting conditions, the precision of our method is quite high; it means there are a few false positives and in the majority of cases the detected object is a kiwifruit. 
    
    Processing time is one of the most important indicators for a kiwifruit detector. Fu et al. \cite{Fu2015,Fu2017} was the only researcher who released that the processing time for his proposed detection methods was 1.5 and 0.5 second per fruit. In our method adding the preprocessing technique has a negligible effect on the processing time; whereas it has a considerable influence on F$_1$score. In our method, different kiwifruit densities from 53 to 82 kiwifruit per image were tested. The average processing time was $~$1.5 s per image, and it is suitable for using on a real-time robot.
    
    One of the benefits of using FCN-8s was the requirement of a small training dataset and achieving an acceptable result. In our method, training dataset includes 48 images with 200$\times$ 200 pixels and 113 kiwifruits in total. Fu et al. \cite{Fu2018a} trained a Faster R-CNN with 1176 sub-images with 784$\times$784 pixels and 10 kiwifruits on average per image. It means around 11760 kiwifruits were labelled to train the model which is 100 times more than the number of kiwifruits trained in our model. Fu et al. achieved recall and precision of 96.7\% and 89.3\%, respectively compared to our method on occlusion dataset which was 0.75 and 0.92, respectively. Due to the low number of training set our method can detect less kiwifruit than trained faster R-CNN but our method is more reliable. 
    
    \section{Conclusion and Future Work}
    We have presented a semantic segmentation method for detecting kiwifruit in the orchard. The method has been evaluated in different lighting conditions and fruit occlusion based on F$_1$score and processing time. A preprocessing approach is proposed for different lighting conditions which improves the performance. It is shown there are more overexposed pixels in the blue channel of glare images so this channel is replaced with the green channel. On overexposed images and glare images, histogram equalisation is applied to decrease the dynamic lighting conditions. In glare images, the intensity has changed dynamically, so the histogram equalisation is applied on each sub-image, while in overexposed images HE is used on the entire image. We defined occluded fruit, overexposed images and glare images and made a dataset for each definition. The average F$_1$score on images is 0.82 with a processing time of $~$1.5 s per image. However, the performance on glare images was poor, and the proposed preprocessing method increase F$_1$score from 01.3 to 0.42. 
    
    Future work will be adding more sensors to collect the dataset. Integrating the colour camera with a depth camera and training the model with RGBD data can cause better results in typical conditions. The false positives appears to happen most around wires. This can potentially be decreased if wire detection is added. In order to improve the performance of detection it is worthwhile to add more data with: kiwifruit occlusions, titled kiwifruit and various lighting conditions to the training data set.

\section*{Acknowledgments}
    This research was funded by the New Zealand Ministry for Business, Innovation and Employment (MBIE) on contract UOAX1414. 


\begin{thebibliography}{}
        
        \bibitem[{A. R. Jim{\'{e}}nez} et~al., 2000]{A.R.Jimenez2000}
        {A. R. Jim{\'{e}}nez}, {R. Ceres}, and {J. L. Pons} (2000).
        \newblock {A SURVEY OF COMPUTER VISION METHODS FOR LOCATING FRUIT ON TREES}.
        \newblock {\em Transactions of the ASAE}.
        
        \bibitem[Baeten et~al., 2008]{Baeten2008}
        Baeten, J., Donn{\'{e}}, K., Boedrij, S., Beckers, W., and Claesen, E. (2008).
        \newblock {Autonomous fruit picking machine: A robotic apple harvester}.
        \newblock In {\em Springer Tracts in Advanced Robotics}.
        
        \bibitem[Bargoti and Underwood, 2016]{Bargoti2016}
        Bargoti, S. and Underwood, J. (2016).
        \newblock {Image classification with orchard metadata}.
        \newblock In {\em Proceedings - IEEE International Conference on Robotics and
          Automation}.
        
        \bibitem[Bargoti and Underwood, 2017]{Bargoti2017a}
        Bargoti, S. and Underwood, J.~P. (2017).
        \newblock {Image Segmentation for Fruit Detection and Yield Estimation in Apple
          Orchards}.
        \newblock {\em Journal of Field Robotics}, 34.6:1039--1060.
        
        \bibitem[Barnett et~al., 2017]{Barnett2017}
        Barnett, J., Seabright, M., Williams, H., Nejati, M., Scarfe, A., Bell, J.,
          Jones, M., Martinson, P., and Schare, P. (2017).
        \newblock {Robotic Pollination - Targeting Kiwifruit Flowers for Commercial
          Application}.
        \newblock {\em International Tri-Conference for Precision Agriculture}.
        
        \bibitem[Brown and K, 1968]{Brown1968}
        Brown, C. E.~S. and K, G. (1968).
        \newblock {Basic Considerations in Mechanizing Citrus Harvest}.
        \newblock {\em Transactions of the ASAE}, 11(3):343.
        
        \bibitem[Dias et~al., 2018]{Dias2018}
        Dias, P.~A., Tabb, A., and Medeiros, H. (2018).
        \newblock {Apple flower detection using deep convolutional networks}.
        \newblock {\em Computers in Industry}, 99:17--28.
        
        \bibitem[Dobrusin et~al., 1992]{Dobrusin1992}
        Dobrusin, Y., Edan, Y., Grinshpun, J., Peiper, U.~M., and Hetzroni, A. (1992).
        \newblock {Real-time image processing for robotic melon harvesting}.
        
        \bibitem[Fu et~al., 2018]{Fu2018a}
        Fu, L., Feng, Y., Majeed, Y., Zhang, X., Zhang, J., Karkee, M., and Zhang, Q.
          (2018).
        \newblock {Kiwifruit detection in field images using Faster R-CNN with ZFNet}.
        \newblock {\em IFAC-PapersOnLine}, 51(17):45--50.
        
        \bibitem[Fu et~al., 2017]{Fu2017}
        Fu, L., Sun, S., Manuel, V.~A., Li, S., Li, R., and Cui, Y. (2017).
        \newblock {Kiwifruit recognition method at night based on fruit calyx image}.
        \newblock {\em Nongye Gongcheng Xuebao/Transactions of the Chinese Society of
          Agricultural Engineering}.
        
        \bibitem[Fu et~al., 2015]{Fu2015}
        Fu, L.~S., Wang, B., Cui, Y.~J., Su, S., Gejima, Y., and Kobayashi, T. (2015).
        \newblock {Kiwifruit recognition at nighttime using artificial lighting based
          on machine vision}.
        \newblock {\em International Journal of Agricultural and Biological
          Engineering}.
        
        \bibitem[Gongal et~al., 2015]{Gongal2015}
        Gongal, A., Amatya, S., Karkee, M., Zhang, Q., and Lewis, K. (2015).
        \newblock {Sensors and systems for fruit detection and localization: A review}.
        \newblock {\em Computers and Electronics in Agriculture}, 116(July):8--19.
        
        \bibitem[Hiroma, 2002]{Hiroma2002}
        Hiroma, T. (2002).
        \newblock {A Color Model for Recognition of Apples by a Robotic Harvesting
          System}.
        \newblock {\em JOURNAL of the JAPANESE SOCIETY of AGRICULTURAL MACHINERY},
          64(5):123--133.
        
        \bibitem[Li et~al., 2011]{Li2011}
        Li, P., Lee, S.-H., and Hsu, H.-Y. (2011).
        \newblock {Procedia Engineering Review on fruit harvesting method for potential
          use of automatic fruit harvesting systems}.
        \newblock {\em Procedia Engineering}, 23:351--366.
        
        \bibitem[Long et~al., 2015]{Long2015}
        Long, J., Shelhamer, E., and Darrell, T. (2015).
        \newblock {Fully Convolutional Networks for Semantic Segmentation}.
        \newblock In {\em IEEE Transactions on Pattern Analysis and Machine
          Intelligence}, pages 3431--3440.
        
        \bibitem[Nejati et~al., 2019]{Nejati2019}
        Nejati, M., Ahn, H.~S., and MacDonald, B. (2019).
        \newblock {Design of a sensing module for a kiwifruit flower pollinator robot}.
        \newblock {\em Australasian Conference on Robotics and Automation, ACRA},
          2019-December.
        
        \bibitem[Nguyen et~al., 2016]{Nguyen2016}
        Nguyen, T.~T., Vandevoorde, K., Wouters, N., Kayacan, E., {De Baerdemaeker},
          J.~G., and Saeys, W. (2016).
        \newblock {Detection of red and bicoloured apples on tree with an RGB-D
          camera}.
        \newblock {\em Biosystems Engineering}.
        
        \bibitem[Robinson and Assignment, 1955]{Robinson1955}
        Robinson, J. and Assignment, S. (1955).
        \newblock 83-97, 1955.
        \newblock {\em Naval research logistics quarterly}.
        
        \bibitem[Sa et~al., 2016]{Sa2016a}
        Sa, I., Ge, Z., Dayoub, F., Upcroft, B., Perez, T., and McCool, C. (2016).
        \newblock {DeepFruits: A Fruit Detection System Using Deep Neural Networks}.
        \newblock {\em Sensors}.
        
        \bibitem[Safety, 2016]{Safety2016}
        Safety, W. (2016).
        \newblock {for WORKPLACE SAFETY PROTECTION FROM EYE INJURY}.
        \newblock Technical report.
        
        \bibitem[Silwal et~al., 2017]{Silwal2017}
        Silwal, A., Davidson, J.~R., Karkee, M., Mo, C., Zhang, Q., and Lewis, K.
          (2017).
        \newblock {Design, integration, and field evaluation of a robotic apple
          harvester}.
        \newblock {\em Journal of Field Robotics}.
        
        \bibitem[Simonyan and Zisserman, 2014]{Simonyan2014}
        Simonyan, K. and Zisserman, A. (2014).
        \newblock {Very Deep Convolutional Networks for Large-Scale Image Recognition}.
        \newblock pages 1--14.
        
        \bibitem[Wang et~al., 2013]{Wanga2013}
        Wang, Q., Nuske, S., Bergerman, M., and Singh, S. (2013).
        \newblock {Automated Crop Yield Estimation for Apple Orchards}.
        \newblock {\em Experimental robotics}, pages 745--758.
        
        \bibitem[Wijethunga et~al., 2009]{Wijethunga2009}
        Wijethunga, P., Samarasinghe, S., Kulasiri, D., and Woodhead, I. (2009).
        \newblock {Towards a generalized colour image segmentation for kiwifruit
          detection}.
        \newblock In {\em 2009 24th International Conference Image and Vision Computing
          New Zealand, IVCNZ 2009 - Conference Proceedings}.
        
        \bibitem[Williams et~al., 2019a]{Williams2019}
        Williams, H., Nejati, M., Hussein, S., Penhall, N., Lim, J.~Y., Jones, M.~H.,
          Bell, J., Ahn, H.~S., Bradley, S., Schaare, P., Martinsen, P., Alomar, M.,
          Patel, P., Seabright, M., Duke, M., Scarfe, A., and MacDonald, B. (2019a).
        \newblock {Autonomous pollination of individual kiwifruit flowers: Toward a
          robotic kiwifruit pollinator}.
        \newblock {\em Journal of Field Robotics}, (August 2018).
        
        \bibitem[Williams et~al., 2019b]{Williams2019H2}
        Williams, H., Ting, C., Nejati, M., Jones, M.~H., Penhall, N., Lim, J.,
          Seabright, M., Bell, J., Ahn, H.~S., Scarfe, A., Duke, M., and MacDonald, B.
          (2019b).
        \newblock {Improvements to and large‐scale evaluation of a robotic kiwifruit
          harvester}.
        \newblock {\em Journal of Field Robotics}.
        
        \bibitem[Williams et~al., 2019c]{Williams2019H1}
        Williams, H.~A., Jones, M.~H., Nejati, M., Seabright, M.~J., Bell, J., Penhall,
          N.~D., Barnett, J.~J., Duke, M.~D., Scarfe, A.~J., Ahn, H.~S., Lim, J.~Y.,
          and MacDonald, B.~A. (2019c).
        \newblock {Robotic kiwifruit harvesting using machine vision, convolutional
          neural networks, and robotic arms}.
        \newblock {\em Biosystems Engineering}, 181:140--156.
        
        \bibitem[Zemmour et~al., 2017]{Zemmour2017}
        Zemmour, E., Kurtser, P., and Edan, Y. (2017).
        \newblock {Dynamic thresholding algorithm for robotic apple detection}.
        \newblock In {\em 2017 IEEE International Conference on Autonomous Robot
          Systems and Competitions, ICARSC 2017}.
        
        \bibitem[{Zhan, W., He, D., {\&} Shi}, 2013]{ZhanW}
        {Zhan, W., He, D., {\&} Shi}, S. (2013).
        \newblock {[2013]-TCSAE-C7-Recognition of kiwifruit in field based on Adaboost
          algorithm}.
        \newblock {\em Transactions of the Chinese Society of Agricultural
          Engineering}, 29(23):140--146.
        
        \bibitem[Zhao et~al., 2016]{Zhao2016}
        Zhao, Y., Gong, L., Huang, Y., and Liu, C. (2016).
        \newblock {A review of key techniques of vision-based control for harvesting
          robot}.
        \newblock {\em Computers and Electronics in Agriculture}, 127:311--323.
        
    \end{thebibliography}
    
\end{document}